\documentclass[a4paper,twoside]{article}
\usepackage{epsfig}
\usepackage{subcaption}
\usepackage{calc}
\usepackage{amssymb}
\usepackage{amstext}
\usepackage{amsmath}
\usepackage{amsthm}
\usepackage{multicol}
\usepackage{pslatex}
\usepackage{apalike}
\usepackage{tikz}
\usepackage{algorithm2e}
\usepackage[bottom]{footmisc}
\usepackage{SCITEPRESS} 

\begin{document}

\title{Towards Ubiquitous Mapping and Localization \\for Dynamic Indoor Environments}

\author{\authorname{Halim Djerroud\sup{1}, Nico Steyn\sup{2}, Olivier Rabreau\sup{1}, Patrick Bonnin\sup{1}, Abderraouf Benali\sup{1}}
\affiliation{\sup{1}Université Paris-Saclay, UVSQ, LISV, 78140, Vélizy-Villacoublay, France.}
\affiliation{\sup{2}Tshwane University of Technology. Department of Electrical Engineering, Pretoria, South Africa}
\email{\{halm.djerroud, olivier.rabreau, patrick.bonnin, abderraouf.benali\}@uvsq.fr, steynn@tut.ac.za}
}


\abstract{
We present UbiSLAM, an innovative solution for real-time mapping and localization in dynamic indoor environments. By deploying a network of fixed RGB-D cameras strategically throughout the workspace, UbiSLAM addresses limitations commonly encountered in traditional SLAM systems, such as sensitivity to environmental changes and reliance on mobile unit sensors. This fixed-sensor approach enables real-time, comprehensive mapping, enhancing the localization accuracy and responsiveness of robots operating within the environment. The centralized map generated by UbiSLAM is continuously updated, providing robots with an accurate global view, which improves navigation, minimizes collisions, and facilitates smoother human-robot interactions in shared spaces. Beyond its advantages, UbiSLAM faces challenges, particularly in ensuring complete spatial coverage and managing blind spots, which necessitate data integration from the robots themselves. In this paper we discusse a potential solutions, such as automatic calibration for optimal camera placement and orientation, along with enhanced communication protocols for real-time data sharing. The proposed model reduces the computational load on individual robotic units, allowing less complex robotic platforms to operate effectively while enhancing the robustness of the overall system.
}

\onecolumn \maketitle \normalsize \setcounter{footnote}{0} \vfill

\section{\uppercase{Introduction}}
\label{sec:introduction}
Traditional SLAM (Simultaneous Localization and Mapping) systems \cite{taheri2021slam}, though well-established, reveal significant limitations \cite{huang_critique_2016} in dynamic environments where human-robot interactions are frequent. In SLAM, sensors are directly mounted on the robots (often LiDAR, RGB-D cameras, and IMUs), meaning that the map and localization are incrementally built as the robot explores the environment. This makes SLAM highly dependent on the detection capabilities of each mobile robot and vulnerable to rapid environmental changes, such as obstacles movements or the arrival of humans in the mapped space \cite{saputra2018visual}. These disruptions can introduce inconsistencies and cumulative errors (drift, sensor errors, etc.) in the generated maps, making map updates and obstacles positions less reliable.

\begin{figure*}[!h]
  \centering
  \includegraphics[width = 14.5cm]{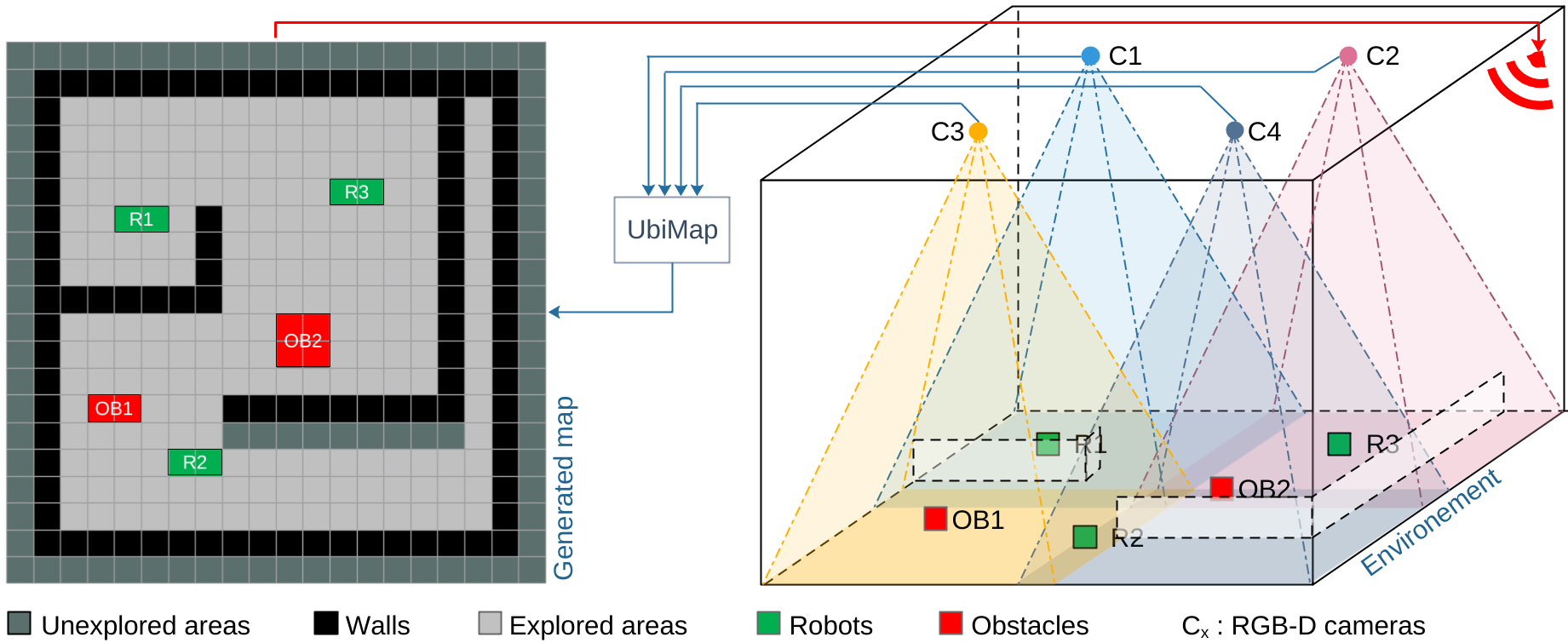}
  \caption{Real-time mapping system of an indoor environment with multiple robots (R1, R2, R3) and obstacles (OB1, OB2). 
  On the right: a 3D representation of the RGB-D sensor network (C1, C2, C3, C4) positioned to cover different areas.
  On the left: a generated map displays the exploration status of the environment. Explored areas are shown in light gray, unexplored areas in dark gray, and walls are represented in black. Robots are marked in green (R1, R2, R3) and obstacles in red (OB1, OB2). The map produced is then transmitted to the various robots via a wireless link.}
  \label{fig:ubislam}
\end{figure*}

In this article, we present an innovative approach based on the use of fixed sensors within the environment to overcome the limitations of traditional SLAM methods, which we will refer to as \textit{UbiSLAM} (Ubiquitous SLAM). This method utilizes a network of strategically positioned RGB-D cameras to capture and generate a real-time, detailed map of the environment. This setup enables precise localization of both obstacles and robots within the space (see Figure \ref{fig:ubislam}). The resulting map is then transmitted to the various robots operating within this environment.

However, \textit{UbiSLAM} relies on a more complex initial deployment and calibration of the fixed sensor network, particularly to avoid blind spots and minimize redundancy in coverage. Nonetheless, for applications in dynamic indoor environments, \textit{UbiSLAM} represents a substantial advancement over traditional SLAM, offering more robust, accurate, and interoperable real-time mapping.

The remainder of this article is structured as follows: the article begin with a review of existing SLAM methods and the associated challenges in dynamic environments. Next, the detail of the architecture of \textit{UbiSLAM} and the specifics of its deployment in real-world settings are discussed. Then we present the advantages of our approach over conventional methods. Finally, we discuss potential improvements and potential applications of \textit{UbiSLAM}.

\section{\uppercase{Limitations of SLAM in Indoor Environments}}
SLAM (Simultaneous Localization and Mapping) \cite{taheri2021slam} is a fundamental method in mobile robotics and perception that aims to address the challenge of localization and mapping in unknown environments. The core challenge of SLAM lies in the circular dependency between accurate localization and the construction of a coherent map: a robot needs to know its relative pose (position and orientation) to generate an accurate map, yet conversely, it requires a map to accurately determine its pose.

In the literature, two primary approaches emerge to address the SLAM problem. The first, based on the robot’s own movements, is commonly known as Odometry-SLAM \cite{yang2020review}. This method relies on the robot’s model to estimate its current position $x_t$ relative to its initial pose $x_0$, using sensors such as odometers and accelerometers. However, a major drawback of this approach is drift, which affects the accuracy of measurements over time due to the inherent limitations of the sensors and the slip occurring between the robot’s wheels and the ground.

The second main category, known as Visual SLAM \cite{kazerouni2022survey}, uses external elements with sensors like monocular or stereo cameras, LiDAR, or RGB-D cameras. This method relies on environmental features used as landmarks to estimate the robot’s pose. One of the main challenges of this technique lies in the ability to identify the same landmarks between two successive moments, $x_{t-1}$ and $x_t$, to ensure consistent tracking of the position, especially in dynamic environments \cite{nielsen2022multi}. Although Visual SLAM is also prone to drift, this can potentially be corrected by detecting the initial point $x_0$ at the end of the trajectory, this challenge is known as the ``loop closure problem'' \cite{coutsias2004kinematic,chen2021overlapnet}.

Finally, a third category of SLAM is less-developed and is based on using acoustic elements in the environment. This technique, called Acoustic SLAM, leverages an array of microphones to achieve localization based on the acoustic properties of the environment \cite{hu2023acoustic}. This approach is particularly emerging in underwater robotics and is less commonly applied in indoor environments.

Technically, SLAM combines the two mentioned above techniques, often referred to as visual odometry \cite{kazerouni2022survey}. It uses onboard sensors, such as LiDAR, RGB-D cameras, and inertial measurement units (IMUs), which provide sensory information about the environment. Estimation algorithms \cite{strasdat2012visual}, such as the Extended Kalman Filter (EKF-SLAM), particle filters (FastSLAM) \cite{montemerlo2002fastslam}, and graph-based optimization methods (Graph-SLAM) \cite{thrun2006graph}, are used to solve nonlinear state equations and optimize both the robot’s pose and the map of surrounding features. These methods enable error minimization and the handling of sensor uncertainties by Leveraging-Bayesian probability techniques and recursive filtering \cite{nadiga2019leveraging}.

The mathematical formulation of SLAM is based on a probabilistic model, where the objective is to maximize the joint probability of the robot’s state and the map based on observations and movements \cite{kudriashov2020slam}. This problem is generally framed as a sequential Bayesian filtering problem. The goal is to jointly estimate the robot's position $x_t$ and the environment map $m$ using observations $z_t$ and motion commands $u_t$. This process can be described in three steps: (1) prediction, (2) update, and (3) joint estimation (Bayesian filtering) \cite{do2016fully}.

\paragraph{a. Position Prediction:}
The robot estimates its current position $x_t$ based on its previous position $x_{t-1}$ and the applied motion command $u_t$:
\begin{equation}
    p(x_t | x_{t-1}, u_t)
\end{equation}

\paragraph{b. Observation Update:}
The observations $z_t$ are used to adjust the robot's estimated position and update the map $m$ by comparing the current observations with the existing map:
\begin{equation}
    p(z_t | x_t, m)
\end{equation}

The estimation can be performed in several different ways, including:

\paragraph{c.1 Bayesian Filtering}
The position and map are re-estimated by combining predictions with new observations to obtain a more accurate estimate:

\begin{multline}
    p(x_t, m | z_{1:t}, u_{1:t}) = \eta \, p(z_t | x_t, m) \\
    \int p(x_t | x_{t-1}, u_t) p(x_{t-1}, m | z_{1:t-1}, u_{1:t-1}) \, dx_{t-1}
\end{multline}
where $ \eta $ is a normalization factor.

The pose and map estimation is sequential, carried out by incorporating new data progressively to build a map and locate the robot within that map. An erroneous estimate at $t_{-n}$ will accumulate across all estimates required for each $p_{t_-n}$ to $p_{t}$.

\paragraph{c.2 Extended Kalman Filter Approach (EKF-SLAM):}
When the motion and observation models are approximately linear around an estimate, the Extended Kalman Filter (EKF-SLAM) can be used to solve the problem \cite{bailey2006consistency}. This approach approximates the models by linearizing around the current estimate, allowing the position and map estimation to be updated using Gaussian distributions in two steps:

\paragraph{State Prediction:}

   \begin{equation}
       x_{t} = f(x_{t-1}, u_t) + w_t
   \end{equation}
   Where $ w_t \sim \mathcal{N}(0, Q) $ represents model noise.

\paragraph{Observation :} 
   \begin{equation}
       z_t = h(x_t, m) + v_t
   \end{equation}
   Where $ v_t \sim \mathcal{N}(0, R) $ represents observation noise.\\~

These equations and distributions allow for calculating and maintaining a real-time estimate of both pose and the map. SLAM thus becomes a recursive estimation of the map and localization based on sensor data and movement commands within a Bayesian framework. There are numerous variants that attempt to optimize this technique  \cite{zhang2018comparison}.

\subsection{Review of SLAM Methods}
In the following section, we will explore some of the most advanced SLAM methods, including MonoSLAM, PTAM, and ORB-SLAM \cite{taketomi2017visual}. These approaches have played a fundamental role in the development of simultaneous localization and mapping, each offering its own advantages and disadvantages. We will examine them in the chronological order of their emergence.

\subsubsection{MonoSLAM}
MonoSLAM is one of the first implementations of monocular SLAM \cite{davison2003real,davison2007monoslam}. Based on an Extended Kalman Filter (EKF), this method enables the simultaneous estimation of camera motion and 3D structure in an unknown environment. MonoSLAM uses feature points in the image for tracking, with the Kalman filter maintaining a continuous state estimation. However, the method is limited by its computational cost, which increases proportionally with the size of the environment, making real-time application in large environments challenging.

\subsubsection{Parallel Tracking and Mapping}
Parallel Tracking and Mapping (PTAM) was developed to overcome some of the limitations of MonoSLAM by separating camera tracking and mapping into two parallel processes. Introduced by \cite{klein2007parallel}, PTAM uses Bundle Adjustment (BA) optimization to enhance map accuracy. The method relies on keyframes to reconstruct the map, and localization algorithms allow the recovery of the camera’s position if it is lost. Thanks to its multithreaded approach, PTAM can handle thousands of feature points, making it a robust solution for confined indoor environments with limited size.

\subsubsection{ORB-SLAM}
Developed in 2015 \cite{mur2015orb}, ORB-SLAM represents a significant advancement by utilizing robust ORB (Oriented FAST and Rotated BRIEF) features. This open-source and modular method includes modules for relocalization and loop closure, making the system highly stable and accurate even in unstructured environments. ORB-SLAM employs pose graph optimization and global bundle adjustment to maintain the geometric consistency of the map. It supports monocular, stereo, and RGB-D SLAM, making it one of the most comprehensive and versatile SLAM methods. ORB-SLAM ensures precise camera tracking and high-quality mapping, and has been widely adopted in robotics and augmented reality applications due to its open-source nature and ability to run in real-time with limited computational resources.

In 2016, ORB-SLAM2 was developed \cite{mur2017orb} as an extension of ORB-SLAM, supporting not only monocular cameras but also stereo and RGB-D cameras. This version enhances the system’s robustness by leveraging the depth information provided by stereo or RGB-D sensors, allowing it to resolve the scale ambiguities inherent in monocular configurations. ORB-SLAM2 also integrates improvements in loop detection and global map optimization, making the system better suited for dynamic and large-scale environments. By incorporating depth information, ORB-SLAM2 can provide more accurate mapping and improve stability during rapid camera movements.

The latest version, ORB-SLAM3, released in 2021 \cite{campos2021orb}, is the most advanced iteration of the ORB-SLAM series, adding support for visual-inertial fusion. This version allows the system to utilize inertial sensor data (IMU) to improve tracking accuracy in scenarios involving rapid movements or challenging environments for visual feature detection. ORB-SLAM3 supports monocular, stereo, RGB-D, and visual-inertial configurations, making it highly versatile. With enhanced modules for initialization and optimization, as well as for loop detection in complex environments, ORB-SLAM3 is one of the most comprehensive SLAM systems, suitable for advanced robotics applications.

ORB-SLAM, with its successive evolutions, offers a robust and versatile suite of tools capable of adapting to various types of sensors and applications, establishing itself as a standard in visual SLAM. Each of these methods has made a significant contribution to the field of visual SLAM, addressing diverse needs and meeting the challenges of localization and mapping in complex environments.

\subsection{Limitations of SLAM Methods}
Traditional SLAM methods \cite{siciliano_simultaneous_2016}, although well-established, reveal significant limitations in dynamic \cite{xing2022slam} and unstructured environments where frequent human-robot interactions occur \cite{soares2021crowd}. In SLAM, the sensors (often LiDAR, RGB-D cameras, and IMUs) are mounted directly on the robots, meaning that the map and localization are incrementally built as the robot explores the environment. This makes SLAM highly dependent on the detection capabilities of each mobile unit and vulnerable to rapid environmental changes, such as obstacles movements or the presence of humans in the mapped space. These disruptions can introduce inconsistencies and errors in the generated maps, making position updates less reliable. We have analyzed and identified a major issues inherent to SLAM in dynamic environments:

\textbf{Error Accumulation and Drift :} Over time, localization errors accumulate in traditional SLAM systems, causing a phenomenon known as "drift", which distorts the map and compromises positional accuracy. This issue is especially significant in large environments or those lacking fixed reference points, where drift becomes a major problem that requires complex and often computationally expensive correction techniques.

\textbf{Dependence on Onboard Sensors:} As previously mentioned, SLAM methods rely on sensors mounted directly on the robot. This dependence means that mapping and localization are tightly linked to the perception and detection capabilities of the moving robot. Consequently, if a robot loses visual contact or encounters obstacles outside of its detection range, the map’s accuracy may degrade. While advanced methods like ORB-SLAM can correct map drift through loop closure by detecting the initial point $x_0$ at the end of a path, their effectiveness is largely confined to relatively static environments. In settings where obstacles frequently shift positions—such as those with moving humans or other robots—these methods often face inconsistencies in obstacle localization, as real-time updates become increasingly challenging in a continuously changing space.

\textbf{Scalability Issues :} As the environment to be mapped grows, computational and memory demands increase, making it challenging to extend traditional SLAM to large or complex environments without performance degradation. Systems must maintain an increasingly large and detailed map, which can slow down real-time processing.

\textbf{Perception Limits and Computational Complexity :} Due to the nonlinear complexity of simultaneously estimating pose and mapping, SLAM systems require robust and computationally intensive algorithms \cite{samsuri2015computational}, which can make challenges for implementation on robots with limited computational capabilities. Algorithms like the Extended Kalman Filter (EKF) or particle-based techniques are resource-demanding, limiting their real-time application on low-cost robots or those with minimal processing power.

\textbf{Lack of Coordination in Multi-Robot Systems :} In applications where multiple robots work together \cite{chen2023overview}, each robot often handles its own localization and mapping, which can lead to discrepancies between the maps generated by each unit. Traditional SLAM approaches lack built-in mechanisms for map fusion or synchronization of localization information across multiple robots, limiting their collaborative capability.

The primary difference with a ubiquitous approach, which we will now refer to as \textit{UbiSLAM}, lies in the fact that the sensors are fixed and provide continuous coverage of the entire operational area. Rather than relying on a mobile unit to build and update the map, \textit{UbiSLAM} leverages a network of RGB-D cameras strategically placed in the environment, creating a real-time map accessible to all mobile units. This approach overcomes the fragmented and dynamic nature of traditional SLAM mapping. Indeed, the fixed sensor network maintains a comprehensive and up-to-date view of the space, making the map less susceptible to sudden movements and configuration changes.

\textit{UbiSLAM} also offers advantages in computational cost and reliability. By centralizing mapping and reducing the need for each robot to perform simultaneous localization and mapping, the system allows for the use of less powerful robots while still ensuring accurate mapping. Furthermore, the integration of human posture detection within UbiSLAM enhances safety and interaction in applications where humans and robots coexist, in contrast to traditional SLAM systems that do not naturally integrate this type of tracking.

\section{\uppercase{Ubiquitous Mapping Approach}}

The proposed mapping approach, UbiSLAM, utilizes a network of fixed sensors -- primarily RGB-D cameras -- strategically positioned throughout the environment to enable robust and precise mapping independent of the sensors onboard the robots. This paradigm aims to overcome the limitations of traditional SLAM systems by eliminating the cumulative error sources inherent to onboard sensors and enhancing system resilience to rapid environmental changes.

RGB-D cameras are installed in optimized locations to ensure complete coverage of the workspace. These cameras provide a stable data source, allowing precise tracking of the positions of robots and obstacles within the environment. The cameras calculated strategic distributed and oriented positions are done in order to minimize blind spots and optimally maximize observational redundancy, enabling robust and continuous monitoring across the environment. This configuration not only mitigates the risk of coverage gaps due to possible partial sensor network failures, but also addresses potential latency issues by allowing overlapping fields of view. In cases of network delay or sensor lag, the overlapping coverage ensures that critical areas are still monitored in near real-time, as adjacent cameras can supplement delayed or missing data.

Data collected by each camera is centralized and fused in real time, enabling the production of a global map of the environment with enhanced accuracy. This approach can use multi-sensor fusion techniques. The generated map is continuously updated to reflect changes in the environment (such as objects movements or the appearance of new obstacles) and is made accessible to all robots operating within this space.

Each robot uses the centralized map provided by the system, rather than relying solely on its onboard sensors for localization. By integrating this information, robots benefit from continuous localization correction, reducing cumulative errors and enhancing movement accuracy. This shared map also enables better coordination among robots, promoting multi-robot collaboration without collision risks and facilitating information sharing in shared environments.

By offloading the mapping and localization processing, UbiSLAM significantly reduces the computational load and energy consumption of the robots. With centralized data processing, robots can operate with lighter, energy-efficient hardware, improving their autonomy and adaptability in environments where extended operational duration is essential.

The use of fixed RGB-D cameras ensures greater resilience to frequent disturbances and dynamic changes in the environment. Robots can adapt to new spatial configurations in real time without losing precision or efficiency. Furthermore, the redundancy of observations from fixed cameras enhances the system’s robustness against temporary occlusions and rapid environmental changes.

In conclusion, the ubiquitous mapping approach proposed by UbiSLAM represents a significant advancement over traditional SLAM methods, providing a collaborative, distributed, and highly resilient mapping solution. UbiSLAM will bring notable improvements in localization accuracy and map quality, paving the way for new applications in collaborative robotics, service robotics, and indoor industrial automation.

\subsection{Challenges Inherent to \textit{UbiSLAM}:}

The proposed solution raises several issues that must be addressed to ensure a high-performance solution. A few key challenges have been identified which are:

\paragraph{a. Optimization of Camera Placement:}
The optimal positioning of RGB-D cameras is crucial to achieve maximum coverage of the space while minimizing uncovered areas and excessive redundancies. Camera placement must be calculated to cover strategic zones where robots and humans frequently interact. This requires determining each camera’s viewing angle, range, and height, taking into account spatial constraints and potential obstacles that might block the field of view. A rectangular tiling optimization approach can be employed to ensure complete coverage of the space while minimizing the number of cameras required.

\paragraph{b. Automatic Camera Calibration:}
Initial calibration and ongoing maintenance of fixed cameras represent a major challenge, as each camera must be precisely aligned with the global coordinate system to ensure consistent mapping. This includes adjusting orientations, correcting optical distortions\footnote{The issue of optical distortions is not addressed further in this article.}, and determining the exact position of each camera within the environment. Automatic calibration would simplify the initial deployment of the sensor network and facilitate reconfiguration if the camera layout changes.

\paragraph{c. Real-Time Communication Protocol:}
A robust, low-latency communication system is essential for enabling the instantaneous sharing of localization and mapping data between fixed sensors and mobile robots. The protocol must ensure data synchronization to provide a consistent view of the environment at all times. In environments with potential blind spots, the global map can be supplemented with data from robot sensors and then merged. Data fusion from robot sensors requires a bidirectional communication protocol.

\paragraph{d. Merging Maps Generated by Robots:}
In collaborative environments, each robot may generate its own local map based on information collected by its onboard sensors \cite{liu2023efficient}. Merging these local maps with the global map \cite{cristofalo2020geod}, created by the network of fixed cameras, is essential for ensuring consistency in navigation and localization across robots. This fusion requires sophisticated data processing algorithms capable of integrating maps while correcting potential errors from inconsistencies or redundancies between the local and global maps. A consensus system is necessary for this purpose \cite{gao2020random}. Additionally, this fusion process must be adaptable to frequent environmental changes, such as obstacles movement or spatial reconfigurations.

In summary, the challenges associated with implementing \textit{UbiSLAM} highlight the complexity of a collaborative mapping and localization system in a dynamic environment. Optimizing camera placement and calibration, the need for a real-time communication protocol, and the merging of local and global maps are all issues that require robust solutions to ensure optimal accuracy and adaptability to environmental changes. In the following sections, we will outline potential solutions for the first two challenges.

\section{\uppercase{Camera Placement Optimization}}
To optimize the placement of RGB-D cameras, we propose using a rectangular tiling technique \cite{allauzen1997tiling}. In this approach, the space is divided into regular square cells, each representing a unit of the area to be covered. For such tiling, each camera is positioned so that its field of view covers a group of square cells, thereby maximizing coverage with minimal overlap. We consider the environment as a discrete space divided into a regular grid of cells $S = \{ s_1, s_2, \dots, s_n \}$ where each $s_j \in S$  is a region or cell within the space that needs coverage, with each cell representing a potential coverage point for a sensor. 

The RGB-D cameras placed in the environment have a limited field of view, defined by a viewing angle and maximum range, and we assume that all cameras have the same characteristics. Let $C=\{c_1,c_2,...,c_n\}$ represent the set of cameras, where each camera $c_i$ has a field of view $FOV_i$ determined by its coverage angle and maximum range.

We define a coverage function $f:C\rightarrow 2^s$ that associates each camera position $c_i$ with the set of cells $s_j$ covered by $c_i$.

$$                 
f(c_i)=\{s_j \in S : \text{the sensor placed at } c_i \text{ can cover } s_j\}
$$

The coverage range is defined by the camera's field of view and maximum distance. The total coverage of the environment by a set of cameras $P$ is therefore:

$$
\text{Total coverage}(P) = \bigcup_{c_i \in P} f(c_i)
$$

The objective is to optimize the subset $P \subseteq C$ to maximize the total coverage of $S$, minimizing uncovered areas, increasing redundancy in strategic areas (such as zones where robots frequently interact), and minimizing the number of cameras for cost and complexity reasons. This can be formulated as an integer linear programming problem \cite{wolsey2020integer}.

To optimize the placement of RGB-D cameras while considering height above the ground, we define the following parameters based on the 3D projections of the field of view onto the ground plane.

\paragraph{a. Camera Height:} Let $ h_i $  represent the height of camera $ c_i $ above the ground. This height influences both the range and shape of the covered area on the ground based on projection geometry. It is also important to note that the height at which the camera is positioned can affect data quality \cite{rodriguez2021comparison}.

\paragraph{b. Ground Projection of Field of View:}
Consider the field of view of a camera defined by a vertical coverage angle $\beta$ and a horizontal coverage angle $\alpha$. The maximum projected range on the ground, $ d_i $, for camera $ c_i $ is given by:
$$
   d_i = h_i \cdot \tan\left(\frac{\beta}{2}\right)
$$
Where $ h_i $ is the height of the camera. This allows us to define the rectangular coverage projected onto the ground, with a maximum depth $ d_i $ and a width $ w_i = 2 \cdot h_i \cdot \tan\left(\frac{\alpha}{2}\right) $, where $ \alpha $ is the horizontal field of view angle.

\paragraph{c. Ground Coverage Area:} The projection of each camera $c_i$'s, coverage area, positioned at $(x_i, y_i)$ on the ground plane, takes the form of a rectangle with dimensions $ w_i \times d_i $, oriented along the camera's direction. Thus, a point $(x, y)$ is covered by $ c_i $ if, after transformation by the camera's orientation angle $\theta$, it satisfies :

$$
   x_i - \frac{w_i}{2} \leq x \leq x_i + \frac{w_i}{2}
$$
   And
$$
   y_i \leq y \leq y_i + d_i.
$$

\paragraph{d. Coverage Function:} 
Let $ S = \{ s_1, s_2, \dots, s_n \} $ be the set of ground cells to cover, where each cell $ s_j \in S $ represents a discrete region of the space. The total coverage function, which we aim to maximize, is given by:

$$
   f(C, S) = \sum_{s_j \in S} \min \left(1, \sum_{c_i \in C} \text{proj\_cov}_{c_i}(s_j) \right)
$$
   Where $ \text{proj\_cov}_{c_i}(s_j) $ is an indicator function equal to 1 if cell $ s_j $ falls within the ground projection of the field of view of $ c_i $, and 0 otherwise.

\paragraph{e. Overlap Optimization:}
To minimize coverage redundancies while ensuring complete coverage of $S$, we introduce an overlap constraint:

$$
   m \leq \sum_{c_i \in C} \text{proj\_cov}_{c_i}(s_j) \leq k
$$

For each cell $ s_j \in S $, where $ k $ is the maximum allowable overlap threshold and $m$ étant le seuil de chevauchement minimal. is the minimum overlap threshold. It is important to maintain a non-zero space between $m$ and $k$ to ensure an overlap zone where landmarks can be placed for sensor calibration. The following section addresses this issue in more details.

This modeling provides a rigorous mathematical framework for an optimization problem that combines camera height, coverage angles, and ground projections to maximize the coverage efficiency of the space $S$. It is important to note that $S$ can be limited strictly to critical areas to reduce the number of cameras required, leaving shadowed zones in areas deemed non-critical. This perspective is not covered by the technique described above.

\section{\uppercase{Cameras on the Same Plane}}
Determining the position of each camera relative to a global reference $R_0$ is crucial for calibrating multiple RGB-D cameras, enabling consistent object reconstruction and precise data fusion to generate a detailed map. In the proposed system, multiple cameras are positioned to cover the space with a slight overlap $m$ between adjacent fields of view. This small overlap is utilized to align coordinate systems using landmarks visible to adjacent sensors. In each overlap zone, easily detectable landmarks are placed, visible in the RGB-D data of both cameras involved, thus serving as common references for aligning their local coordinate systems. These landmarks can take the form of visual markers integrated into the environment, such as color markers, or distinct-shaped objects that are easily identifiable in depth data.

For each pair of cameras sharing an overlap zone, the coordinates of the landmarks are recorded in each camera's local coordinate system. The RGB-D camera captures both color and depth data, thereby \textit{precisely} locating the landmarks in 3D within the local spaces of the cameras. These landmark coordinates are then used to estimate the geometric transformation (rotation and translation) between the coordinate systems of adjacent cameras, applying algorithms such as ICP (Iterative Closest Point) \cite{zhang2021fast} or correspondence-based methods. These techniques minimize the error between corresponding landmarks, providing a reliable estimation of the transformation.

Once transformations are calculated for each camera pair, they are integrated into a transformation graph \cite{sooriyaarachchi2022elastic}, where each node represents a camera, and each edge represents the estimated transformation between two adjacent cameras. Starting from a global reference frame $R_0$, the transformations are propagated through the graph to align each camera to this global reference. To minimize accumulated errors during propagation, a global optimization is applied to the cameras’ positions and orientations, employing SLAM techniques to reduce calibration errors. This optimization enhances accuracy by adjusting transformations to maintain the overall coherence of the system.

\begin{tikzpicture}
\fill[blue!10] (0,0) -- (-2,-2) -- (2,-2) -- cycle; 
\fill[green!10] (3,0) -- (1,-2) -- (5,-2) -- cycle; 
\fill[orange!40, opacity=0.8] (1,-1.5) rectangle (2,-2);

\node[draw, rectangle, fill=blue!20] (C1) at (0,0) {$C_1$};
\node[draw, rectangle, fill=green!20] (C2) at (3,0) {$C_2$};

\draw[->] (0,0) -- (-0.5,0.5) node[above] {$z_1$};
\draw[->] (3,0) -- (3.5,0.5) node[above] {$z_2$};

\filldraw[red] (1.2,-1.8) circle (1.5pt) node[above] {$p_1$};
\filldraw[red] (1.8,-1.9) circle (1.5pt) node[above] {$p_2$};
\filldraw[red] (1.5,-1.6) circle (1.5pt) node[above] {$p_3$};

\node at (0,-1) [blue] {Field of View $C_1$};
\node at (3,-1) [green] {Field of View $C_2$};
\node at (1.5,-2.3) [orange] {Overlap Zone};
\end{tikzpicture}

To calibrate RGB-D cameras in a three-dimensional space using overlap zones, we propose a system that uses ICP \cite{wang2017survey,chetverikov2002trimmed} to estimate rigid transformations between cameras sharing common landmarks in the overlap zones. The ICP algorithm is particularly suited to this situation, as it minimizes the error between two point sets by adjusting the rotation and translation of one set relative to the other.
First, rigid transformations $T_{ij}$ are defined for each pair of cameras that share an overlap zone where landmarks are identified. Then, ICP is applied to estimate these transformations by minimizing the distances between corresponding landmarks. A transformation graph is subsequently constructed to link the cameras, allowing for transformation propagation within a global reference frame $R_0$, thereby expressing each camera within this global frame. Finally, a global optimization is performed to reduce accumulated errors and ensure precise calibration across the entire system.

\paragraph{Rigid Transformation with ICP:}
For each pair of cameras with an overlap zone, we define a rigid transformation $T_{ij}$ that aligns the coordinate system of camera $i$ with that of camera $j$. This transformation $T_{ij}$ includes a rotation matrix $R_{ij}$ and a translation vector $t_{ij}$ :
$$
T_{ij} = \begin{bmatrix} R_{ij} & t_{ij} \\ 0 & 1 \end{bmatrix}
$$
Where $ R_{ij} \in \mathbb{R}^{3 \times 3} $ and $ t_{ij} \in \mathbb{R}^3 $.

\paragraph{Acquisition of Landmark Points in Overlap Zones:}
For each pair of cameras sharing an overlap zone, we obtain two sets of landmark points: $P_i = \{p_i^1, p_i^2, \ldots, p_i^n\} $ in the coordinate system of camera $i $ and $P_j = \{p_j^1, p_j^2, \ldots, p_j^n\} $ in the coordinate system of camera $j$. These landmark points serve as inputs for the ICP algorithm.

\paragraph{Application of the ICP Algorithm:}
The ICP algorithm is applied to align points $P_i$ from camera $i$ to points $P_j$ from camera $j$, minimizing the distance between each pair of corresponding points. ICP performs the following steps:
(1)\textbf{Matching Step} : For each point in $P_i$, the closest point in $P_j$ is identified.
(2)\textbf{Optimization Step} : The transformation $ T_{ij} $ (comprising $ R_{ij} $ and $ t_{ij} $) is calculated to minimize the sum of squared distances between corresponding points. This optimization is solved using the least squares method.
These steps are repeated until convergence is reached, meaning that the average difference between the distances of corresponding points falls below a specified threshold $s$.

\paragraph{Construction du Graphe de Transformations:}
Once the transformations $T_{ij}$ are calculated for each camera pair, the cameras are modeled as a graph where each camera is a node, and each transformation $T_{ij}$ represents an edge connecting two adjacent cameras.

\paragraph{Propagation of Transformations to the Global Reference Frame $R_0$:}
To express each camera in the global reference frame $R_0$, we multiply transformations along the paths in the graph. For example, if camera $k$ is connected to $R_0$ via cameras $i$ and $j$, then the global transformation $T_{0k}$ is given by:

$$
T_{0k} = T_{0i} T_{ij} T_{jk}
$$

\paragraph{Global Transformation Optimization:}
To minimize accumulation errors in the graph, a global optimization is applied by minimizing a cost function $ C $, which represents the sum of matching errors for each camera pair. The cost function is defined by:

$$
C = \sum_{(i,j)} \sum_{k} \| T_{ij} p_i^k - p_j^k \|^2
$$

Where $ p_i^k $ and $ p_j^k $ are the coordinates of the $ k $-th landmark in the coordinate systems of cameras $i$ and $j$. This optimization adjusts the transformations $ T_{ij} $ to minimize the total calibration error in the system. Optimization methods such as Levenberg-Marquardt (LM algorithm) \cite{ranganathan2004levenberg} will be used to refine the transformations and minimize inconsistencies.

In this section, we proposed an ICP-based approach that enables precise calibration of RGB-D cameras by using overlap zones to align each camera within a global reference frame $R_0$. This alignment allows all entities in the environment to orient themselves within a common reference frame, thus simplifying localization challenges.

\section{\uppercase{Localization}}

To reliably locate the robots, each robot is equipped with a unique visual identifier, easily detectable by RGB-D cameras positioned throughout the space. This identifier, placed on top of each robot, provides a precise landmark that enables the system to calculate the robot’s exact position in the global reference frame $R_0$. The cameras continuously capture the identifier code, and the system deduces the robot's spatial coordinates based on the relative position of the code with respect to the cameras.

Additionally, for obstacles localization and human detection, a technology such as the YOLO object detection model \cite{diwan2023object} can be integrated into the system. This allows real-time object detection and identification of moving objects, including people, and situates them in space relative to $R_0$. The model proposed here functions effectively in dynamic environments, where it can identify unexpected obstacles or humans entering the workspace. A fixed localization method utilizing a large network of RGB-D cameras offers several advantages in dynamic environments such as dense, collaborative manufacturing setups. By deploying these cameras throughout the space, UbiSLAM enables precise tracking of robots, static objects, and human movement. This approach ensures accurate and continuous localization without relying on onboard sensors, offering enhanced robustness in complex environments where both environmental dynamics and operational efficiency are crucial.

One other key benefit of this fixed localization method is its ability to track essential specialized items stored in less frequently used areas of the manufacturing process. In large, busy manufacturing warehouses, locating goods or assets that are not in regular use can be a significant challenge, often resulting in inefficiencies and delays. The comprehensive data provided by the RGB-D camera network allows for consistent monitoring of these assets, ensuring they are not only tracked in real-time but also stored and retrieved efficiently when needed. By creating a digital map of the environment that continuously updates and monitors key items, the system reduces the risk of misplaced or under-utilized resources.

Additionally, the ability to store and access historical data on asset location is particularly valuable for optimizing workflows and inventory management. Over time, this data can be leveraged to predict patterns, improve operational strategies, and streamline storage practices. While this approach still faces challenges in large-scale applications, such as dealing with sensor network failures or latency issues, the potential of UbiSLAM to enhance the visibility and traceability of goods and assets in manufacturing environments is promising. Future exploration of UbiSLAM's application in these contexts could further validate its effectiveness in improving inventory tracking and asset management, ultimately benefiting large-scale, collaborative manufacturing systems.

\section{\uppercase{Discussion}}
We discussed the advantages of having a fixed mapping and localization system. However, this system is subject to certain limitations, notably areas not covered by the camera network, commonly called blind spots. To address these, data from the robots must be integrated, adding a layer of complexity to interoperability between \textit{UbiSLAM} and the robots.

The map is generated in real time and continuously updated, integrating environmental changes such as obstacles movements and the presence of people. This map is then shared with the robots navigating within the space, providing them with an up-to-date global view of the environment. Robots can thereby access location information for obstacles and other agents, enhancing navigation and reducing collision risks. Through this centralized mapping approach, the map generated by \textit{UbiSLAM} is not only precise but also adaptable in real time to environmental dynamics, facilitating collaboration between robots and humans in shared spaces.

The \textit{UbiSLAM} approach represents significant advances in mapping and localization for dynamic indoor environments, particularly where human-robot interactions are frequent and unpredictable. In contrast to traditional SLAM methods, \textit{UbiSLAM} employs a network of fixed sensors that provides real-time, accurate, and comprehensive mapping of the environment, thereby reducing the cumulative localization errors typical of systems based on onboard sensors. Strategically placed RGB-D cameras enable continuous spatial modeling that is less affected by rapid environmental changes. This architecture reduces the computational load on mobile robots by externalizing mapping, leading to better energy management and allowing simpler robotic units to operate reliably. Moreover, the considerable reduction in RGB-D camera costs makes this solution easily applicable in small to medium-sized indoor environments. The price of a 3D LiDAR mounted on a robot can be up to 20 times that of an RGB-D sensor.

The mathematical methods used in \textit{UbiSLAM}, such as multi-sensor data fusion, can be extended with probabilistic models like Bayesian filtering and the Extended Kalman Filter (EKF), to handle uncertainties associated with sensor measurements.

However, challenges remain, especially in achieving complete environmental coverage. Blind spots, or areas without coverage, can lead to localization loss for robots and require rigorous optimization of sensor placement. Modeling this aspect through discrete coverage techniques, where each camera covers spatial units according to a coverage function $f(C)$, could maximize coverage while minimizing excessive overlap, thus reducing deployment costs and complexity. The resulting combinatorial optimization problem is often addressed using heuristics such as sensor placement optimization algorithms, which aim to find an optimal distribution of cameras in the space but typically yield approximate solutions.

The integration of real-time detection models, such as YOLO for identifying obstacles and people, offers a valuable complement to identifier code-based robot localization. YOLO enables rapid and reliable recognition of moving elements, making \textit{UbiSLAM} suitable for detecting and tracking moving objects without adding additional sensors to the robots. This framework presents promising prospects for collaborative multi-robot environments, necessitating further research to efficiently coordinate multiple units and improve methods for merging local and global maps. Future developments in \textit{UbiSLAM} could focus on network resource management to accelerate data flow and apply advanced data fusion techniques to further enhance the accuracy and robustness of the system in constantly changing environmental conditions.

\section{\uppercase{Conclusion}}

The proposed \textit{UbiSLAM} approach marks a significant advancement in mapping and localization for dynamic indoor environments. By integrating a network of fixed sensors strategically placed throughout the environment, this method overcomes the limitations of traditional SLAM systems, which are often hindered by obstacles changes and an inability to adapt in real time. By providing comprehensive and continuous mapping, \textit{UbiSLAM} enhances localization accuracy and robot responsiveness, enabling improved human-robot interaction in shared spaces. This ubiquitous model reduces the computational load on mobile units, ensuring increased robustness to rapid environmental changes and optimizing the safety and fluidity of interactions.

However, \textit{UbiSLAM} presents certain challenges that merit further attention. Achieving complete spatial coverage remains an issue, particularly in addressing blind spots (areas not covered by fixed sensors), which require solutions to ensure coherent and continuous mapping. Integrating automatic calibration techniques to optimize camera placement and orientation, along with improving communication protocols for real-time information sharing, represent promising development avenues. Finally, extending this approach to more complex environments and evaluating its effectiveness in collaborative multi-robot systems will open up new possibilities for applications in service and industrial robotics.

\textit{UbiSLAM} offers a robust solution for ubiquitous mapping and localization, enhancing navigation and collaboration in complex indoor environments, thereby laying the groundwork for a new generation of intelligent systems for collaborative robotics.

\bibliographystyle{apalike}
{\small
\bibliography{biblio}}

\end{document}